\documentclass[final]{cvpr}

\usepackage{times}
\usepackage{epsfig}
\usepackage{graphicx}
\usepackage{amsmath}
\usepackage{amssymb}
\usepackage{kotex}

\usepackage{multirow}
\usepackage{xcolor}
\usepackage{makecell}
\usepackage{adjustbox}
\usepackage{url}
\usepackage{enumitem}
\usepackage{flushend}
\usepackage{tabularx}
\usepackage{makecell}
\usepackage{booktabs}
\usepackage[nocompress]{cite} 

\usepackage[pagebackref=true,breaklinks=true,colorlinks,bookmarks=false]{hyperref}

\begin{document}

\title{XProtoNet: Diagnosis in Chest Radiography with Global and Local Explanations}

\author{Eunji Kim$^1$ ~~~~~~~ Siwon Kim$^1$ ~~~~~~~  Minji Seo$^1$  ~~~~~~~  Sungroh Yoon$^{1, 2, }$\thanks{Correspondence to: Sungroh Yoon (sryoon@snu.ac.kr).}\\
$^1$ Department of Electrical and Computer Engineering, Seoul National University, Seoul, South Korea\\
$^2$ ASRI, INMC, ISRC, and Institute of Engineering Research, Seoul National University\\
{\tt\small \{kce407, tuslkkk, minjiseo, sryoon\}@snu.ac.kr}}

\maketitle

\newcommand{\xvec}{\mathbf{x}}
\newcommand{\pvec}{\mathbf{p}}
\newcommand{\qvec}{\mathbf{q}}
\newcommand{\pveckc}{\pvec_k^c}
\newcommand{\qveckc}{\qvec_k^c}
\begin{abstract}
Automated diagnosis using deep neural networks in chest radiography can help radiologists detect life-threatening diseases. However, existing methods only provide predictions without accurate explanations, undermining the trustworthiness of the diagnostic methods. Here, we present \mbox{XProtoNet}, a globally and locally interpretable diagnosis framework for chest radiography. \mbox{XProtoNet} learns representative patterns of each disease from X-ray images, which are prototypes, and makes a diagnosis on a given X-ray image based on the patterns. It predicts the area where a sign of the disease is likely to appear and compares the features in the predicted area with the prototypes. It can provide a global explanation, the prototype, and a local explanation, how the prototype contributes to the prediction of a single image. Despite the constraint for interpretability, \mbox{XProtoNet} achieves state-of-the-art classification performance on the public NIH chest X-ray dataset.
\end{abstract}
\begingroup
\hyphenpenalty 500

\section{Introduction}
Chest radiography is the most widely used imaging examination for diagnosing heart and other chest diseases~\cite{kelly2012chest}. Detecting a disease through chest radiography is a challenging task that requires professional knowledge and careful observation. Various automated diagnostic methods have been proposed to reduce the burden placed on radiologists and the likelihood of mistakes; methods using deep neural networks (DNNs) have achieved especially high levels of performance in recent decades~\cite{rajpurkar2017chexnet,wang2017chestx,li2018thoracic,hermoza2020region,irvin2019chexpert}. However, the black-box characteristics of DNNs discourage users from trusting DNN predictions~\cite{miller2019explanation,ching2018opportunities}.
Since medical decisions may have life-or-death consequences, medical-diagnosis applications require not only high performance but also a strong rationale for judgment.
Although many automated diagnostic methods have presented localization as an explanation for prediction~\cite{wang2017chestx,rajpurkar2017chexnet,irvin2019chexpert,taghanaki2019infomask,ma2019multi}, this provides only the region on which the network is focusing within a given image, not the manner by which the network makes a decision~\cite{rudin2019stop}.

\begin{figure}[t]
	\centering
    \includegraphics[width=0.98\columnwidth]{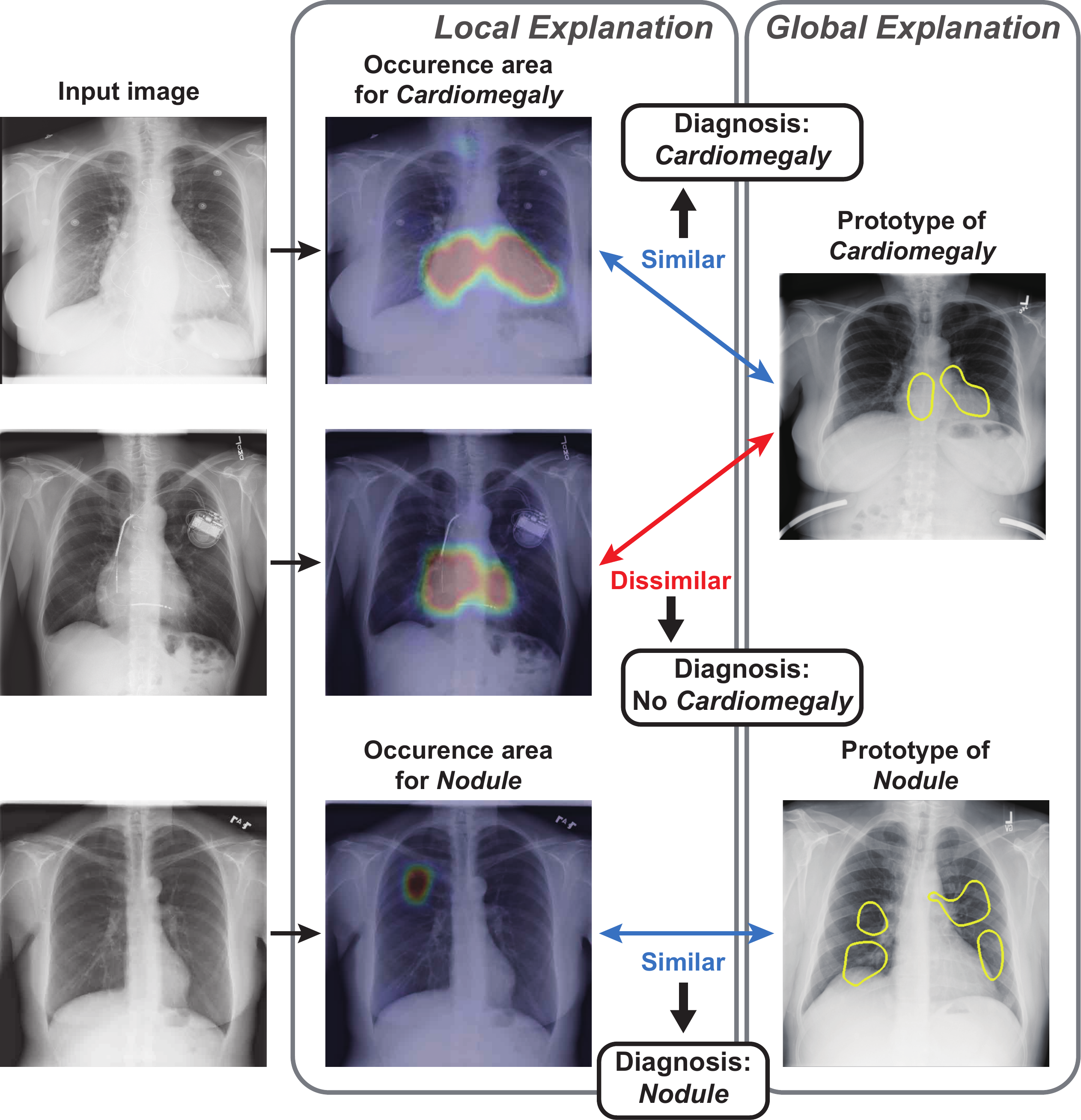}
    \caption{
    Our proposed framework, XProtoNet, learns prototypes that are used to identify each disease.
    Given an input image, \mbox{XProtoNet} compares the feature in the occurrence area of the input image with the prototypes and thereby diagnoses diseases.
    Yellow contours denote the learned prototypes.
    }
    \label{fig_figabstract}
\end{figure}

Interpretable models, unlike conventional neural networks, are designed to operate in a human-understandable manner~\cite{rudin2019stop}. Case-based models learn discriminative features of each class, which are referred to as prototypes, and classify an input image by comparing its features with the prototypes~\cite{li2018deep,chen2019looks,hase2019interpretable}.
Such models provide two types of interpretation: global and local explanations. A global explanation is a class-representative feature that is shared by multiple data points belonging to the same class~\cite{kim2018interpretability,ribeiro2016should}. A local explanation, by contrast, shows how the prediction of a single input image is made.
In other words, the global explanation finds the common characteristic by which the model defines each class, while the local explanation finds the reason that the model sorts a given input image into a certain class.
The global explanation can be likened to the manner in which radiologists explain common signs of diseases in X-ray images, whereas the local explanation can be likened to the manner in which they diagnose individual cases by examining the part of a given X-ray image that provides information about a certain disease. 
This suggests that case-based models are suitable for building an interpretable automated diagnosis system.

ProtoPNet~\cite{chen2019looks}, which motivates our work, defines a prototype as a feature within a patch of a predefined size obtained from training images, and compares a local area in a given input image with the prototypes for classification.
Despite such constraint for interpretability, it achieves performance comparable to that of conventional uninterpretable neural networks in fine-grained classification tasks. However, with a patch of a predefined size, it is difficult to reflect features that appear in a dynamic area, such as a sign of disease in medical images. For example, to identify cardiomegaly (enlargement of the heart), it is necessary to look at the whole heart~\cite{rassi2006development}; to identify nodule, it is necessary to find an abnormal spot whose diameter is smaller than a threshold~\cite{hansell2008fleischner}. Depending on the fixed size of the patch, the prototypes may not sufficiently present the class-representative feature or may even present a class-irrelevant feature, leading to diagnostic failure. To address this problem, we introduce a method of training the prototypes to present class-representative features within a dynamic area (see the prototypes of each disease in Figure~\ref{fig_figabstract}).

In this paper, we propose an interpretable automated diagnosis framework, XProtoNet, that predicts an occurrence area where a sign of a given disease is likely to appear and learns the disease-representative features of the occurrence area as prototypes. The occurrence area is adaptively predicted for each disease, enabling the prototypes to present discriminative features for diagnosis within the adaptive area for the disease.
Given a chest X-ray image, \mbox{XProtoNet} diagnoses disease by comparing the features of the image with the learned prototypes.
As shown in Figure~\ref{fig_figabstract}, it can provide both global explanations—the discriminative features allowing the network to screen for a certain disease—and local ones—\eg, a rationale for classifying a single chest X-ray image.
We evaluate our method on the public NIH chest X-ray dataset~\cite{wang2017chestx}, which provides 14 chest-disease labels and a limited number of bounding box annotations.
We also conduct further analysis of XProtoNet with a prior condition to have specific features as prototypes using the bounding box annotations. Despite strong constraints to make the network interpretable, XProtoNet achieves state-of-the-art diagnostic performance.

The main contributions of this paper can be summarized as follows:
\begin{itemize}
\setlength{\itemsep}{1pt}
	\item[$\bullet$] We present, to the best of our knowledge, the first interpretable model for diagnosis in chest radiography that can provide both global and local explanations.
	\item[$\bullet$] We propose a novel method of learning disease-representative features within a dynamic area, improving both interpretability and diagnostic performance.
	\item[$\bullet$] We demonstrate that our proposed framework outperforms other state-of-the-art methods on the public NIH chest X-ray dataset.
\end{itemize}
\endgroup
\begin{figure*}[ht]
	\centering
    \includegraphics[width=\textwidth]{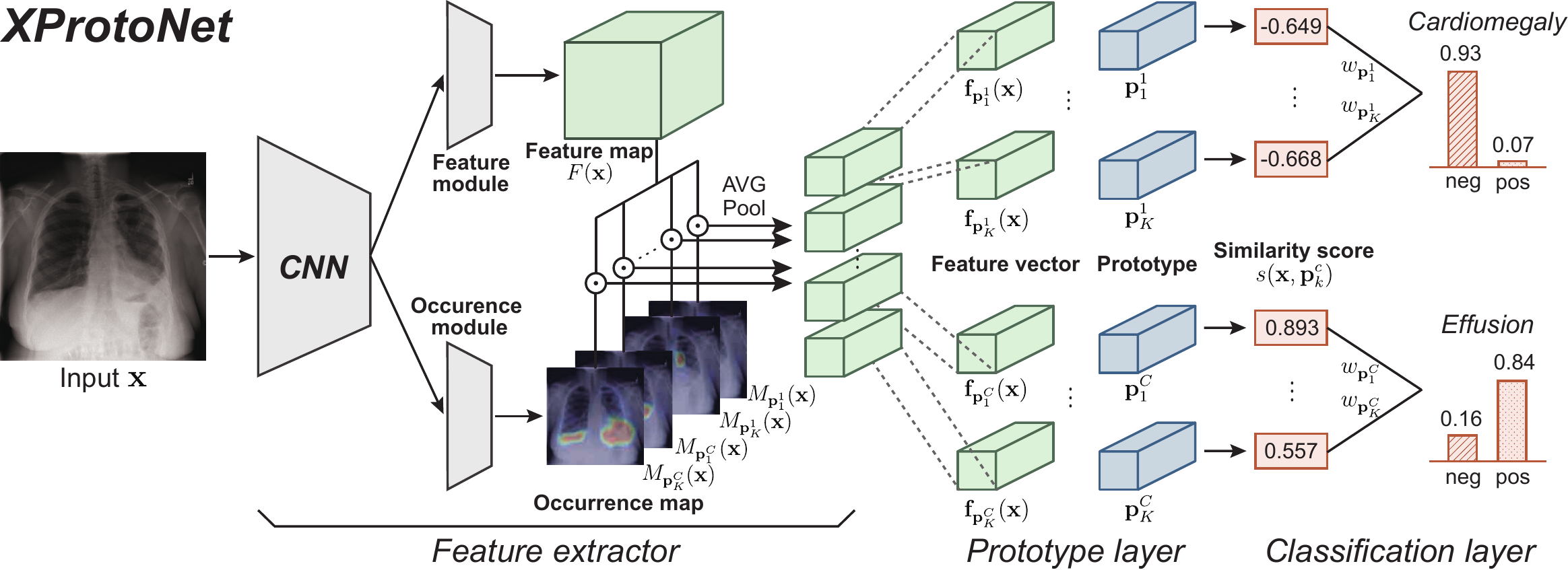}
    \vspace{-1pt}
    \caption{Overall architecture of XProtoNet. XProtoNet diagnoses diseases by comparing the features of an input image to the prototypes of each disease.
    }
    \label{fig:architecture}
\vspace{-1pt}
\end{figure*}
\section{Related Work}
\subsection{Automatic Chest X-ray Analysis}
A number of researchers have attempted to identify diseases via chest radiography using DNNs. Wang~\etal~\cite{wang2017chestx} and Rajpurkar~\etal~\cite{rajpurkar2017chexnet} proposed the use of a conventional convolutional neural network to localize disease through a class activation map~\cite{zhou2016learning}.
Taghanaki~\etal~\cite{taghanaki2019infomask} utilized a variational online mask on a negligible region within the image and predicted disease using the unmasked region.
Guan~\etal~\cite{guan2020multi} proposed a class-specific attention method and Ma~\etal~\cite{ma2019multi} used cross-attention with two conventional convolutional neural networks. Hermoza~\etal~\cite{hermoza2020region} used a feature pyramid network~\cite{lin2017feature} and an additional detection module to detect disease.
Li~\etal~\cite{li2018thoracic} proposed a framework to simultaneously perform disease identification and localization, exploiting a limited amount of additional supervision. Liu~\etal~\cite{liu2019align}, also utilizing additional supervision, proposed a method to align chest X-ray images and learn discriminative features by contrasting positive and negative samples.
Some of these approaches localize the disease along with classification but cannot explain the predictive process of how this localized part contributes to model prediction.
Herein, we aim to build a diagnostic framework to explain the predictive process rather than simply localize the disease.

\subsection{Interpretable Models}
There have been various post-hoc attempts to explain already-trained models~\cite{simonyan2013deep,selvaraju2017grad,smilkov2017smoothgrad,ancona2018towards,kindermans2018learning}, but some of them provide inaccurate explanations~\cite{sixt2019explanations,adebayo2018sanity}. Additionally, they only show the region where the network is looking within a given image~\cite{rudin2019stop}. To address this problem, several models have been proposed with structurally built-in interpretability~\cite{melis2018towards,li2018deep,chen2019looks,hase2019interpretable}.
Since their prediction process itself is interpretable, they require no additional effort to obtain interpretation after training.
A self-explaining neural network~\cite{melis2018towards} obtains both concepts that are crucial in classification and the relevance of each concept separately through regularization, then combines them to make a prediction.
Case-based interpretable models, mostly inspiring us, learn prototypes that present the properties of the corresponding class and identify the similarity of the features of a given input image to the learned prototypes~\cite{li2018deep,chen2019looks,hase2019interpretable}. Li~\etal~\cite{li2018deep} used an encoder-decoder framework to extract features and visualize prototypes. Chen~\etal~\cite{chen2019looks} defined prototypes as a local feature of the image and visualized the prototypes by replacing them with the most similar patches of training data.
Hase~\etal~\cite{hase2019interpretable} proposed training prototypes in a hierarchical structure.
These works targeted classification tasks in general images, and there was no attempt to make an interpretable automated diagnosis framework for chest radiography. To this end, we propose an interpretable diagnosis model for chest radiography that learns disease-representative features within a dynamic area.
\section{XProtoNet}
Figure~\ref{fig:architecture} shows the overall architecture of our proposed framework, XProtoNet: the feature extractor, prototype layer, and classification layer. We describe the diagnostic process of \mbox{XProtoNet} in Section~\ref{sec_process}, and explain in Section~\ref{sec_occurrencemap} how to extract features within a dynamic area. In Section~\ref{sec_training_scheme}, we describe the overall training scheme.

\subsection{Diagnosis Process}\label{sec_process}
\vspace{-1pt}
\mbox{XProtoNet} compares a given input image to learned disease-representative features to diagnose a disease. It has a set of $K$ learned prototypes $\mathcal{P}^c=\{\pveckc\}_{k=1}^K$ for each disease $c$, where the prototype $\pveckc$ presents a discriminative feature of disease $c$. Given an input image $\xvec$, the feature extractor extracts the feature vector $\textbf{f}_{\pveckc}(\xvec)$ for each prototype $\pveckc$, and the prototype layer calculates a similarity score $s$ between $\textbf{f}_{\pveckc}(\xvec)$ and $\pveckc$, which are $\rm{D}$-dimensional vectors.
Similarity score $s$ is calculated using cosine similarity as
\begin{equation}\label{eq:map}
    \vspace{-1pt}
    s(\xvec, \pvec_k^c) = \frac{\textbf{f}_{\pveckc}(\xvec)\cdot\pveckc}{\|\textbf{f}_{\pveckc}(\xvec)\|\|\pveckc\|}.
\end{equation}

Diagnosis from chest radiography is a multi-label classification, which is a binary classification of each class. We thus derive the prediction score of target disease $c$ by considering only the prototypes of $c$, not the prototypes of the non-target diseases, in the classification layer. The prediction score is calculated from
\begin{equation}\label{eq:score_cls2}
    p(y^c|\xvec) = \sigma\left(\sum_{\pvec_k^c\in\mathcal{P}^c}{w_{\pvec_k^c}{s\left(\xvec, \pvec_k^c\right)}}\right),
\end{equation}
where $w_{\pveckc}$ denotes the weight of $\pveckc$ and $\sigma$ represents a sigmoid activation function.
Similarity score $s$ indicates how similar the feature of the input image is to each prototype, and weight $w_{\pveckc}$ indicates how important each prototype is for the diagnosis. By this process, \mbox{XProtoNet} can diagnose the disease based on the similarity between the corresponding prototypes and the features of the input X-ray image.
After the training, prototype $\pveckc$ is replaced with the most similar feature vector $\textbf{f}_{\pveckc}$ from the training images. This enables the prototypes to be visualized as human-interpretable training images, without an additional network for decoding the learned prototype vectors.

\subsection{Extraction of Feature with Occurrence Map}\label{sec_occurrencemap}
\begin{figure}[t]
	\centering
    \includegraphics[width=\columnwidth]{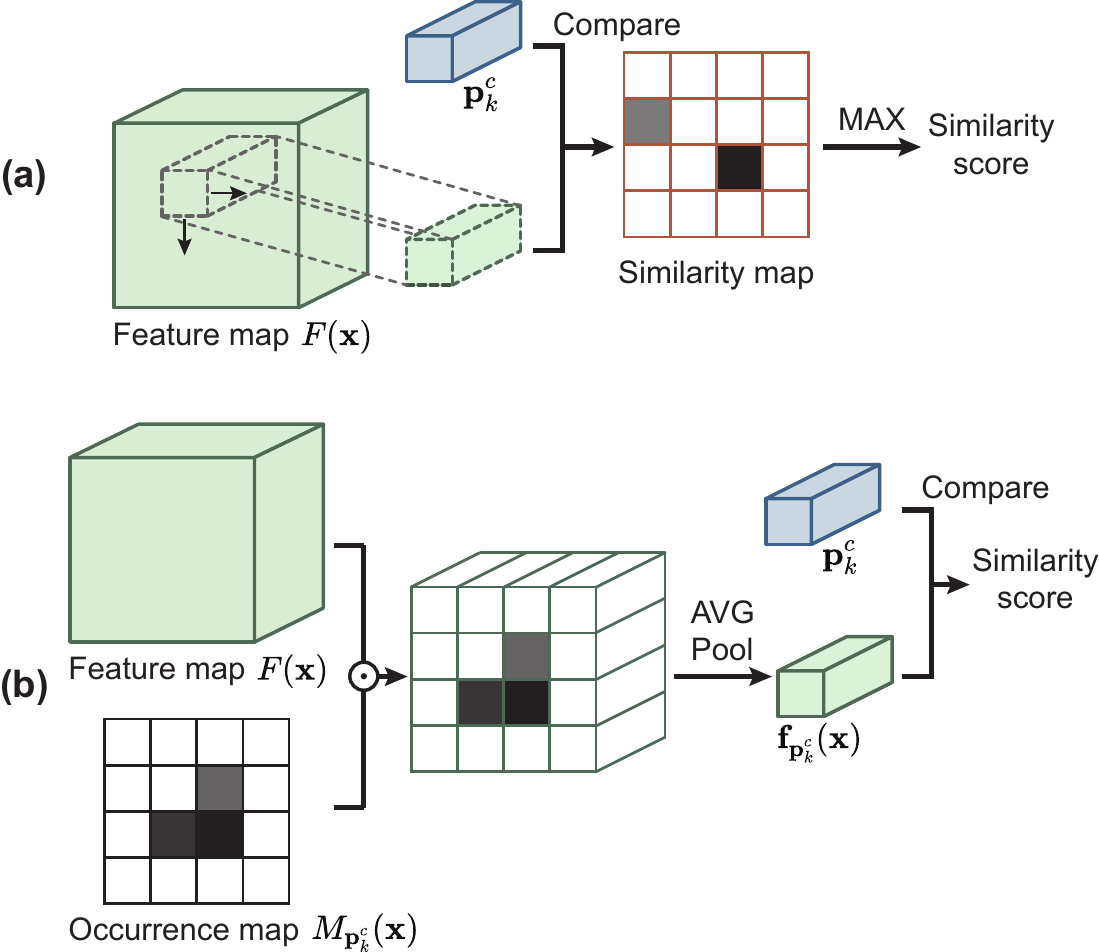}
    \caption{
    Comparison of how XProtoNet and ProtoPNet~\cite{chen2019looks} obtain the similarity of the features of an image with prototype $\pveckc$. Whereas (a) ProtoPNet compares the feature patch from all spatial locations of the feature map with the prototype and outputs the maximum value as the similarity score, (b) XProtoNet makes one feature vector $\textbf{f}_{\pveckc}$ with the occurrence map and compares it with the prototype.
    }
    \label{fig:comparison_calsim}
\vspace{-2pt}
\end{figure}

When extracting feature vectors $\textbf{f}_{\pveckc}$, \mbox{XProtoNet} considers two separate aspects of the input image: the patterns within the image and the area on which to focus to identify a certain disease.
Therefore, the feature extractor of \mbox{XProtoNet} contains a feature module and an occurrence module for each one of the above-mentioned aspects.
The feature module extracts the feature map $F(\xvec)\in\mathbb{R}^{\rm{H} \times \rm{W} \times \rm{D}}$, the latent representations of the input image $\xvec$, where $\rm{H}$, $\rm{W}$, and $\rm{D}$ are the height, width, and dimension, respectively. The occurrence module predicts the occurrence map $M_{\pveckc}(\xvec)\in\mathbb{R}^{\rm{H} \times \rm{W}}$ for each prototype $\pveckc$, which presents where the corresponding prototype is likely to appear, that is, the focus area.
Both modules consist of $1\times1$ convolutional layers.
Using occurrence map $M_{\pveckc}(\xvec)$, feature vector $\textbf{f}_{\pveckc}(\xvec)$ to be compared with prototype $\pveckc$ is obtained as follows:
\begin{equation}\label{eq:occurrence_map}
\vspace{-1pt}
\textbf{f}_{\pveckc}(\xvec)=\sum_{u}{M_{\pveckc,u}(\xvec) F_u(\xvec)},
\vspace{-1pt}
\end{equation}
where $u\in[0, \rm{H}\times\rm{W})$ denotes the spatial location of $M_{\pveckc}(\xvec)$ and $F(\xvec)$ (Figure~\ref{fig:comparison_calsim}(b)). The values of occurrence map, which are in the range $[0,1]$, are used as the weights when pooling the feature map $F(\xvec)$ so that the feature vector $\textbf{f}_{\pveckc}(\xvec)$ represents a feature in the highly activated area in the occurrence map.

By pooling the feature map with the occurrence map, a class-representative feature is presented as a vector of a single size, regardless of the size or shape of the area in which the feature appears. During training, the occurrence area is optimized to cover the area where disease-representative features for each disease appear, and the prototypes become disease-representative features in an adaptive area size. As mentioned in Section~\ref{sec_process}, prototype $\pveckc$ is replaced with the most similar feature vector $\textbf{f}_{\pveckc}$ after training the feature extractor, thus the prototype can be visualized as the occurrence area of the images that the prototype vectors are replaced with.

\textbf{Comparison with ProtoPNet.}
\mbox{XProtoNet} differs from ProtoPNet~\cite{chen2019looks} by being able to learn features within a dynamic area.
In ProtoPNet, the prototypes are compared with fixed-size feature patches from an input image (Figure~\ref{fig:comparison_calsim}(a)).
The spatial size of the prototype is $r\times r$, which is smaller than the feature map.
At all spatial locations in feature map $F(\xvec)$, a patch from $F(\xvec)$ of the same size as prototype $\pveckc$ is compared to the prototype; the maximum value of the resulting similarity map becomes the final similarity score.
Since a fixed-size patch in the feature map is compared with the prototypes, the prototypes can only learn representative patterns within that patch. Thus, the size of the patch greatly affects the classification performance. The prototypes may learn an insufficient portion of the class-representative pattern if the patch is not large enough, and class-irrelevant features may be presented in the prototypes if the patch is too large.
The disease-representative pattern can appear in a wide range of areas, so comparing it with a fixed-size patch may limit the performance.
By contrast, the feature vector in \mbox{XProtoNet} represents the feature throughout the wide range of area predicted by the network, and is not limited to a fixed-size region (Figure~\ref{fig:comparison_calsim}(b)).

\subsection{Training Scheme}\label{sec_training_scheme}
There are four losses in training XProtoNet: classification loss $\mathcal{L}_\text{cls}$, cluster loss $\mathcal{L}_\text{clst}$, separation loss $\mathcal{L}_\text{sep}$, and occurrence loss $\mathcal{L}_\text{occur}$.

\textbf{Classification.}
To address the imbalance in the dataset, a weighted balance loss is used for $\mathcal{L}_\text{cls}$ as in \cite{ma2019multi}:
\begin{equation}\label{loss_cls}
\begin{aligned}
    \mathcal{L}_\text{cls}^c = &-\sum_i{\frac{1}{|N_{\text{pos}}^c|}(1-p^c_i)^\gamma y^c_i}log(p^c_i)\\
    &-\sum_i{\frac{1}{|N_{\text{neg}}^c|}(p^c_i)^\gamma (1-y^c_i)log(1-p^c_i)},
\end{aligned}
\end{equation}
where $p^c_i=p(y^c|\xvec_i)$, the prediction score of the $i$-th sample $\xvec_i$, and $\gamma$ is a parameter for balance. $|N_{\text{neg}}^c|$ and $|N_{\text{pos}}^c|$ denote the number of negative (0) and positive (1) labels on disease $c$, respectively. Further, $y^c_i\in\{0,1\}$ denotes the target label of $\xvec_i$ on disease $c$.

\textbf{Regularization for Interpretability.}
To allow $\pveckc$ to present the characteristics of disease $c$, the similarity between $\xvec$ and $\pveckc$ should be large for a positive sample and small for a negative sample.
Similar to \cite{chen2019looks}, we define cluster loss $\mathcal{L}_\text{clst}$ to maximize the similarity for positive samples and separation loss $\mathcal{L}_\text{sep}$ to minimize the similarity for negative samples:
\begin{equation}\label{eq:loss_clst_sep}
\begin{aligned}
    &\mathcal{L}^c_\text{clst} = -y^c\max_{\pveckc\in \mathcal{P}^c}{s(\xvec,\pveckc)},\\
    &\mathcal{L}^c_\text{sep} = (1-y^c)\max_{\pveckc\in \mathcal{P}^c}{s(\xvec,\pveckc)}.
\end{aligned}
\end{equation}
As in Eq.~\ref{loss_cls}, $\mathcal{L}_\text{clst}^c$ and $\mathcal{L}_\text{sep}^c$ are weighted with the number of negative and positive samples when they are summed over all diseases and samples.

\textbf{Regularization for Occurrence Map.}
To obtain prediction results with good interpretability, it is important to predict an appropriate occurrence map.
Thus, we add two regularization terms to the training of the occurrence module.
As in general object localization~\cite{wang2020self}, since an affine transformation of an image does not change the relative location of a sign of the disease, it should not affect the occurrence map, either.
We thus define the transformation loss $\mathcal{L}^c_\text{trans}$ for disease $c$ as
\begin{equation}\label{eq:loss_trans}
    \mathcal{L}^c_\text{trans} =\sum_{\pveckc\in\mathcal{P}^c}{ \|A(M_{\pveckc}(\xvec))-M_{\pveckc}(A(\xvec))\|_1},
\end{equation}
where $A(\cdot)$ denotes an affine transformation.
We also add $L_1$ loss on the occurrence map to achieve locality of the occurrence area.
It makes the occurrence area as small as possible to avoid covering more regions than necessary.
The occurrence loss $\mathcal{L}_\text{occur}^c$ is thus expressed as
\begin{equation}\label{eq:loss_occur}
    \mathcal{L}_\text{occur}^c = \mathcal{L}_\text{trans}^c + \sum_{\pveckc\in\mathcal{P}^c}{ \|M_{\pvec_k^c}(\xvec)\|_1}.
\end{equation}

\textbf{Overall Cost Function.} All components of the loss are summed over all diseases, so the total loss is expressed as
\begin{equation}\label{eq:loss_total}
    \mathcal{L}_\text{total} =  \mathcal{L}_\text{cls} + \lambda_\text{clst}\mathcal{L}_\text{clst} + \lambda_\text{sep}\mathcal{L}_\text{sep}
    + \lambda_\text{occur}\mathcal{L}_\text{occur},
\end{equation}
where $\lambda_\text{clst}$, $\lambda_\text{sep}$, and $\lambda_\text{occur}$ are hyperparameters for balancing the losses.

\begin{table*}[t]
\setlength{\tabcolsep}{2.4pt}
\centering
\caption{
AUC scores of XProtoNet and various baselines on chest X-ray dataset. The 14 diseases are Atelectasis, Cardiomegaly, Effusion, Infiltration, Mass, Nodule, Pneumonia, Pneumothorax, Consolidation, Edema, Emphysema, Fibrosis, Pleural Thickening, and Hernia, respectively. The name of each disease is shortened to the first four characters (e.g. Atelectasis to Atel). Pne1, Pne2, and P.T. denote Pneumonia, Pneumothorax, and Pleural Thickening, respectively. The term ``w/o $\mathcal{L}_\text{trans}$'' denotes XProtoNet trained without $\mathcal{L}_\text{trans}$.
}
\label{tab:ablation}
\begin{tabular}{l|cccccccccccccc|c}
\Xhline{1pt}
Methods & Atel & Card & Effu & Infi & Mass & Nodu & Pne1 & Pne2 & Cons & Edem & Emph & Fibr & P.T. & Hern & Mean \\
\hline
\hline
Baseline $\text{Patch}_{1\times1}$ &  0.766 &	0.857 &	0.823 &	0.705 &	0.813 &	0.779 &	0.706 &	0.851 &	0.738 &	0.825 &	0.925 &	0.779 &	0.771 &	0.663 &	0.786 \\
Baseline $\text{Patch}_{3\times3}$ &  0.767	& 0.853	& 0.826	& 0.706	& 0.813	& 0.786	& 0.705	& 0.861	& 0.737	& 0.827	& 0.927	& 0.782	& 0.776	& 0.714	& 0.792 \\
Baseline $\text{Patch}_{5\times5}$ &  0.752	& 0.863	& 0.822	& 0.695	& 0.814	& 0.751	& 0.702	& 0.834	& 0.734	& 0.827	& 0.906	& 0.793	& 0.772	& 0.543	& 0.772 \\
Baseline GAP & 0.764 	&0.847 &	0.815 &	0.703 &	0.817 	&0.782 &	0.719 	&0.856 &	0.723 &	0.823 &	0.928 &	0.782 	&0.776 &	0.704 	&0.789 
\\
\hline
XProtoNet (Ours) & \textbf{0.782}	& \textbf{0.881}	& \textbf{0.836}	& \textbf{0.715}	& \textbf{0.834}	& \textbf{0.799}	& \textbf{0.730}	& \textbf{0.874}	& \textbf{0.747}	& \textbf{0.834}	& \textbf{0.936}	& \textbf{0.815}	& \textbf{0.798}	& \textbf{0.896}	& \textbf{0.820} \\
~~w/o $\mathcal{L}_\text{trans}$  & 0.777	& 0.875	& 0.833	& 0.703	& 0.828	& 0.795	& 0.726	& 0.871	& \textbf{0.747}	& 0.832	& 0.934	& 0.806	& 0.796	& 0.892	& 0.815 \\

\Xhline{1pt}
\end{tabular}
\vspace{-4pt}
\end{table*}
\begin{figure*}[t]
	\centering
    \includegraphics[width=0.9\textwidth]{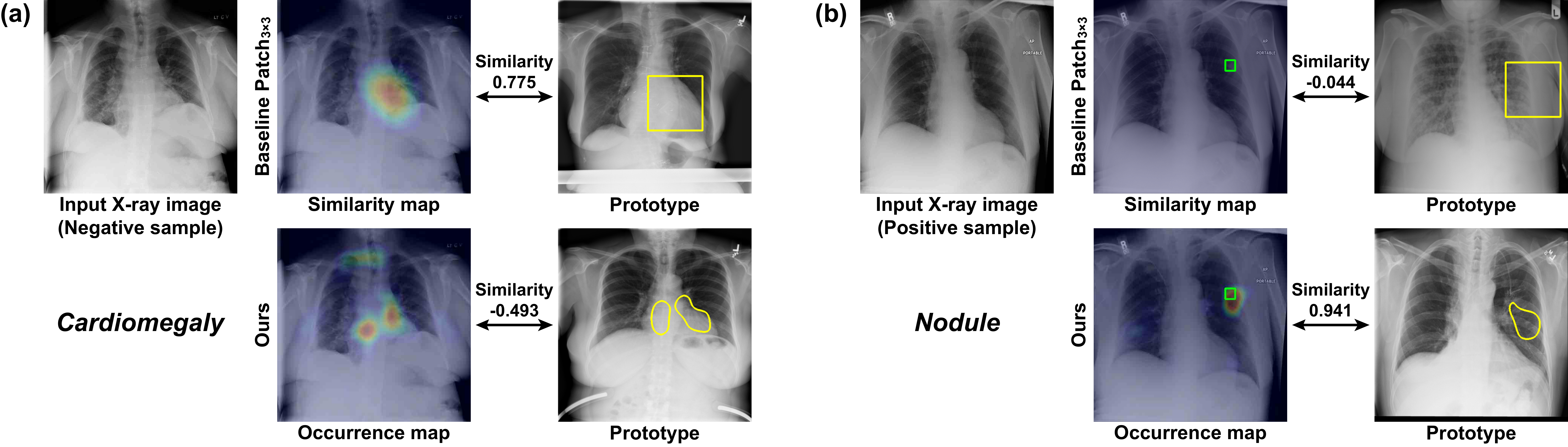}
    \vspace{-1pt}
    \caption{
    Comparison of the predictions between XProtoNet and the baseline $\text{Patch}_{3\times 3}$ for (a) cardiomegaly and (b) nodule diagnoses. The heatmaps are upsampled to the size of the input image. Yellow boxes and contours show the prototypes. Green boxes show the ground-truth bounding boxes from the dataset. There is no bounding box in (a) because it is a negative sample.
    }
    \label{fig:comparesim}
\vspace{-3pt}
\end{figure*}
\begin{figure*}[t]
	\centering
    \includegraphics[width=0.9\textwidth]{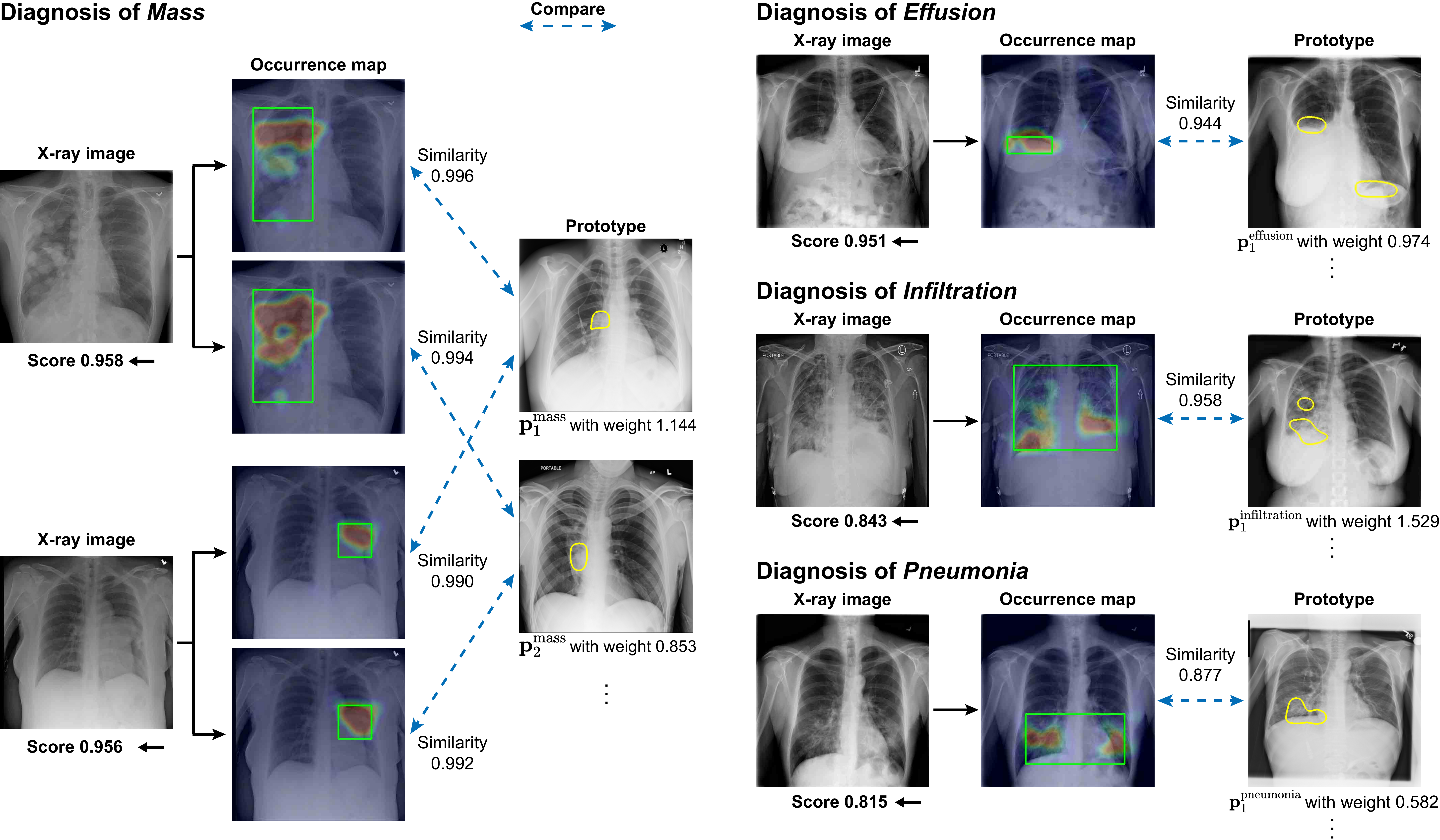}
    \vspace{-1pt}
    \caption{
    Examples of global and local explanations of chest X-rays by XProtoNet. The explanations on one or two input X-ray images are shown with one or two prototypes that have the largest weight on each disease. Yellow contours denote the learned prototypes and green boxes denote the ground truth bounding boxes from the dataset.
    }
    \label{fig:explanation}
\vspace{-5pt}
\end{figure*}

\section{Experiments}

\subsection{Experimental Setup}
\textbf{Dataset.}
The public NIH chest X-ray dataset~\cite{wang2017chestx} consists of 112,120 frontal-view X-ray images with 14 disease labels from 30,805 unique patients. Experiments are conducted with two kinds of data splitting. In most of the experiments, we use an official split that sets aside 20\% of the total images for the test set. We use 70\% for training and 10\% for validation from the remaining images.
In comparison with recent methods using additional supervision (Table~\ref{tab:classification_cv}) and analysis with a prior condition to have specific prototypes (Section~\ref{sec:additional_supervision}), we conduct a five-fold cross validation, similar to that in \cite{li2018thoracic,liu2019align}.
In the official test set, there are 880 images with 984 labeled bounding boxes, provided for only eight types of diseases. We separate the total data into box-annotated and box-unannotated sets and conduct a cross-validation, where each fold has 70\% of each set for training, 10\% for validating, and 20\% for testing. Note that we do not use the bounding box annotation during training, except for analysis with the prior condition.
Patient overlap does not occur between the splits.
We resize images to $512\times512$ and normalize them with ImageNet~\cite{deng2009imagenet} mean and standard deviation. We use data augmentation, by which images are rotated up to $10^{\circ}$ and scaled up or down by 20\% of the image size, similar to that in \cite{hermoza2020region}.

\textbf{Evaluation.}
We evaluate the diagnostic performance of XProtoNet using the area under the receiver operating characteristic curve (AUC) scores.

\textbf{Experimental Details.}
We use ImageNet~\cite{deng2009imagenet} pretrained conventional neural networks as a backbone (\eg, \mbox{ResNet-50}~\cite{he2016deep} and \mbox{DenseNet-121}~\cite{huang2017densely}).
The feature extractor consists of convolutional layers from the backbone network, feature module, and occurrence module. The feature and occurrence modules each consist of two $1\times1$ convolutional layers with ReLU activation between them. The occurrence module has an additional sigmoid activation function to rescale the occurrence value to $[0,1]$. The weights of the classification layer are initially set to 1 so that high similarity scores with the prototypes would result in a high score for the disease.
$K$ and $\rm{D}$ are set to 3 and 128, respectively. The batch size is set to 32. We set $\lambda_{\text{clst}}$, $\lambda_{\text{sep}}$, and $\lambda_{\text{occur}}$ to 0.5. Balance parameter $\gamma$ for $\mathcal{L}_{\text{clst}}$ is set to 2. We use random resizing with ratios 0.75 and 0.875 as affine transformations for $\mathcal{L}_{\text{trans}}$ in Eq.~\ref{eq:loss_trans}.

We follow the training scheme of ProtoPNet~\cite{chen2019looks}: 1) training the model, except for the convolutional layers from the pretrained network and the classification layer, for five epochs; 2) training the feature extractor and the prototype layer until the mean AUC score of the validation set does not improve for three consecutive epochs; 3) replacing the prototypes with the nearest feature vector from the training data; and 4) training the classification layer. The training steps, except for the first step, are repeated until convergence. To retain only supporting prototypes for each disease, prototypes with negative weights are pruned.
More details are explained in the supplementary material.

\textbf{Visualization.}
The occurrence maps are upsampled to the input image size and normalized with the maximum value for visualization.
The prototypes are marked with contours, which depict regions in which the occurrence values are greater than a factor of 0.3 of the maximum value in the occurrence map.

\subsection{Comparison with Baselines}
Table~\ref{tab:ablation} shows the comparison of the diagnostic performance of XProtoNet with various baselines that use different methods of comparison with the prototypes. \mbox{ResNet-50}~\cite{he2016deep} is used as the backbone.
Baseline $\text{Patch}_{r\times r}$ refers to the method that follows ProtoPNet~\cite{chen2019looks} with prototypes of spatial size $r\times r$, as in Figure~\ref{fig:comparison_calsim}(a); baseline GAP refers to the method where the feature vector $\textbf{f}_{\pveckc}(\xvec)$ is obtained by global average pooling (GAP) of the feature map $F(\xvec)$ without an occurrence map. The different performances of the baselines $\text{Patch}_{r\times r}$ show that the performance varies greatly depending on the size of the patch. In addition, the performance of baseline GAP is similar and at times lower than that of baseline $\text{Patch}_{r\times r}$.
By contrast, because XProtoNet predicts the adaptive area to compare, it achieves higher performance in all classes than the baselines: the mean AUC score of $0.820$ is $3.5\%$ higher than the highest baseline $\text{Patch}_{r\times r}$ mean AUC score, which is $0.792$.
Especially, the improvement in hernia is significant ($>25\%$).
This confirms that our proposed method of learning disease-representative features within a dynamic area is effective for diagnosis of medical images. Moreover, $\mathcal{L}_{\text{trans}}$ is also helpful in improving the performance.

Figure~\ref{fig:comparesim} shows the comparison of the predictions between XProtoNet and the baseline $\text{Patch}_{3\times 3}$ which shows the best diagnostic performance among the baselines $\text{Patch}_{r\times r}$.
The cardiomegaly prototype of the baseline presents only a portion of the heart, resulting in a high similarity score (0.775) with the negative sample (Figure~\ref{fig:comparesim}(a)).
By contrast, the prototype of XProtoNet presents almost the whole area of the heart; this is more interpretable than the baseline, and the similarity score between the two occurrence areas is low (-0.369).
Note that the similarity score takes a value in the range $[-1,1]$.
Given the positive sample of nodule (Figure~\ref{fig:comparesim}(b)), XProtoNet successfully detects the small nodule with a high similarity score (0.936) to the prototype, while the baseline fails. In addition, the occurrence area corresponding to the nodule prototype of XProtoNet is consistent with the ground-truth bounding box.
This confirms that our proposed method shows more interpretable visualizations of the prototypes and more accurate predictions than the baseline.

\subsection{Explanation with Prototypes}
Figure~\ref{fig:explanation} shows some examples of the global and local explanations of XProtoNet.
The global explanation of XProtoNet in the diagnosis of mass can be interpreted as follows: the prototypes of mass present an abnormal spot as a major property of mass for XProtoNet; this agrees with the actual sign of lung mass~\cite{hansell2008fleischner}.
In terms of the local explanation of the X-ray image (top left in Figure~\ref{fig:explanation}), \mbox{XProtoNet} predicts that the prototypes of mass are likely to appear in the large left areas of the image, which are consistent with the ground-truth bounding box. XProtoNet outputs high similarity scores between these parts and the corresponding prototypes (0.996 and 0.993), resulting in a high prediction score (0.957) for the mass.
For the diagnosis on the bottom left of Figure~\ref{fig:explanation}, XProtoNet identifies a small region on the right within the image as the occurrence area, which is different from the first example but consistent with the actual sign. This shows that XProtoNet can dynamically predict the appropriate occurrence area.

\begin{figure}[t]
	\centering
    \includegraphics[width=0.88\columnwidth]{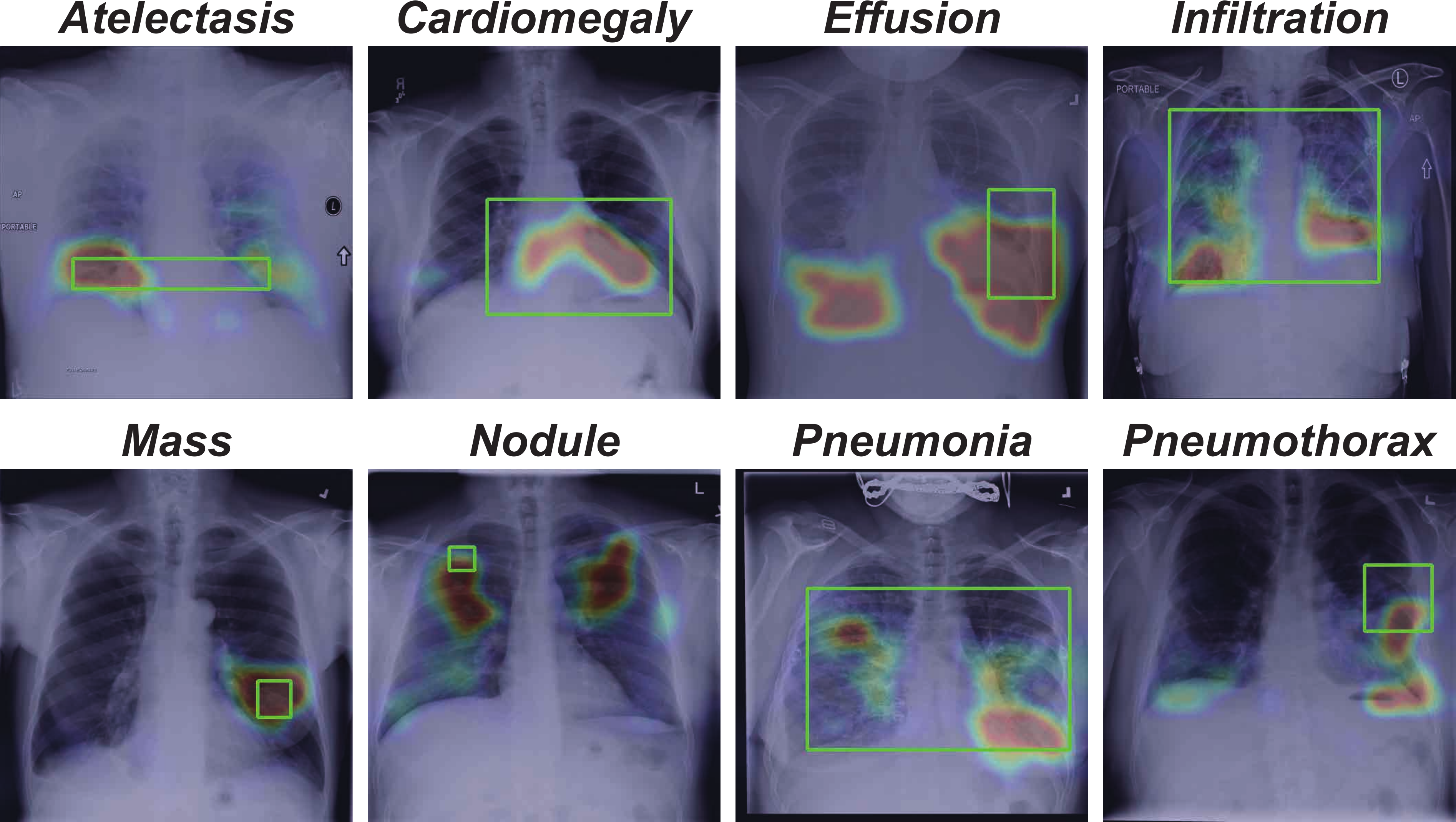}
    \vspace{-1pt}
    \caption{
    X-ray images and occurrence maps that are the most similar to the learned prototypes of each disease. The green boxes are the ground-truth bounding boxes from the dataset.
    }
    \label{fig:occurrence}
\vspace{-5pt}
\end{figure}

To see whether the learned prototypes align with actual signs of diseases, we find the image that is the most similar to the prototype among the images annotated with bounding boxes. 
Note that those annotations are not used during training.
Figure~\ref{fig:occurrence} shows that the occurrence area in the image is consistent with the locus of the actual sign of each disease (green boxes). This shows that the prototypes have been well-trained to present proper disease-representative features.

\subsection{Diagnostic Performance}
We compare the diagnostic performance of XProtoNet with recent automated diagnosis methods~\cite{wang2017chestx,guan2020multi,ma2019multi,hermoza2020region}. Table~\ref{tab:classification} shows that XProtoNet achieves state-of-the-art performance on both \mbox{ResNet-50}~\cite{he2016deep} and \mbox{DenseNet-121}~\cite{huang2017densely} backbones while ensuring interpretability.
In comparison with recent methods implemented on \mbox{ResNet-50}, \mbox{XProtoNet} achieves the best performance for 10 out of 14 diseases.
Note that Ma~\etal~\cite{ma2019multi} use two \mbox{DenseNet-121} and \mbox{Hermoza}~\etal~\cite{hermoza2020region} use a feature pyramid network~\cite{lin2017feature} and \mbox{DenseNet-121} as the backbone: these provide better representation than a single \mbox{DenseNet-121}.
Compared with Guan~\etal~\cite{guan2020multi}, who use a single \mbox{DenseNet-121}, the mean AUC score is improved from $0.816$ to $0.822$.

\begin{table*}[t]
\setlength{\tabcolsep}{2.2pt}
\centering
\caption{AUC scores of XProtoNet and other methods on chest X-ray dataset. The * signifies that an additional conventional network is used as a backbone.
}
\label{tab:classification}
\begin{tabular}{l|cccccccccccccc|c}
\Xhline{1pt}
Methods & Atel & Card & Effu & Infi & Mass & Nodu & Pne1 & Pne2 & Cons & Edem & Emph & Fibr & P.T. & Hern & Mean\\
\hline\hline
\multicolumn{3}{l}{~~Backbone: ResNet-50} \\
Wang~\etal~\cite{wang2017chestx}                          & 0.700                       & 0.810                      & 0.759                     & 0.661                     & 0.693                     & 0.669                     & 0.658                     & 0.799                     & 0.703                     & 0.805                     & 0.833                     & 0.786                     & 0.684                     & 0.872                     & 0.745                        \\
Guan~\etal~\cite{guan2020multi}                                 & 0.779                     & 0.879                     & 0.824                     & 0.694                     & 0.831                      & 0.766                     & 0.726                     & 0.858                     & \textbf{0.758}                     & \textbf{0.850}                     & 0.909                     & \textbf{0.832}                     & 0.778                      & \textbf{0.906}                     & 0.814                        \\
XProtoNet (Ours) & \textbf{0.782}	& \textbf{0.881}	& \textbf{0.836}	& \textbf{0.715}	& \textbf{0.834}	& \textbf{0.799}	& \textbf{0.730}	& \textbf{0.874}	& 0.747	& 0.834	& \textbf{0.936}	& 0.815	& \textbf{0.798}	& 0.896	& \textbf{0.820} \\
\hline
\hline
\multicolumn{6}{l}{~~Backbone: DenseNet-121 / DenseNet-121+$\alpha$*} \\
Guan~\etal~\cite{guan2020multi}                                 & \textbf{0.781}                     & 0.883                     & 0.831                     & 0.697                     & 0.830                      & 0.764                     & 0.725                     & 0.866                     & \textbf{0.758}                     & \textbf{0.853}                     & 0.911                     & 0.826                     & 0.780                      & 0.918                     & 0.816                        \\
Ma~\etal~\cite{ma2019multi}* & {0.777}                     & \textbf{0.894}                     & {0.829}                     & {0.696}                     & \textbf{0.838}                     & {0.771}                     & {0.722}                     & {0.862}                     & {0.750}                      & {0.846}                     & {0.908}                     & {0.827}                     & {0.779}                     & \textbf{0.934}                     & {0.817}                        \\
Hermoza~\etal~\cite{hermoza2020region}* & {0.775}                     & {0.881}                     & {0.831}                     & {0.695}                     & {0.826}                     & {0.789}                     & \textbf{0.741}                     & \textbf{0.879}                     & {0.747}                     & {0.846}                     & {0.936}                     & \textbf{0.833}                     & {0.793}                     & {0.917}                     & 0.821 \\
XProtoNet (Ours) & 0.780	& 0.887	& \textbf{0.835}	& \textbf{0.710}	& 0.831	& \textbf{0.804}	& 0.734	& 0.871	& 0.747	& 0.840	& \textbf{0.941}	& 0.815	& \textbf{0.799}	& 0.909	& \textbf{0.822} \\
\Xhline{1pt}
\end{tabular}
\vspace{-5pt}
\end{table*}
\begin{table*}[t]
\setlength{\tabcolsep}{2.8pt}
\centering
\caption{Comparison with methods that utilize additional bounding box annotations. AUC scores with a five-fold cross-validation performed on the chest X-ray dataset are reported.
Following the previous works, the results are rounded to two decimal digits. The BBox column indicates whether bounding box annotation is used. Note that XProtoNet uses no additional supervision.
}
\label{tab:classification_cv}
\begin{tabular}{l|c|cccccccccccccc|c}
\Xhline{1pt}
Methods & BBox & Atel & Card & Effu & Infi & Mass & Nodu & Pne1 & Pne2 & Cons & Edem & Emph & Fibr & P.T. & Hern & Mean\\
\hline \hline
Li \etal.~\cite{li2018thoracic} & \checkmark & 0.80 & 0.87 & 0.87 & 0.70 & 0.83 & 0.75 & 0.67 & 0.87 & \textbf{0.80} & 0.88 & 0.91 & 0.78 & 0.79 & 0.77 & 0.81\\
Liu \etal.~\cite{liu2019align} & \checkmark & 0.79 & 0.87 & 0.88 & 0.69 & 0.81 & 0.73 & 0.75 & 0.89 & 0.79 & \textbf{0.91} & 0.93 & 0.80 & 0.80 & \textbf{0.92} & 0.83\\
XProtoNet (Ours) & & \textbf{0.83}	& \textbf{0.91}	& \textbf{0.89}	& \textbf{0.72}	& \textbf{0.87}	& \textbf{0.82}	& \textbf{0.76}	& \textbf{0.90}	& \textbf{0.80}	& 0.90	& \textbf{0.94}	& \textbf{0.82}	& \textbf{0.82}	& \textbf{0.92}	& \textbf{0.85} \\
\Xhline{1pt}
\end{tabular}
\vspace{-3pt}
\end{table*}

We also compare the diagnostic performance of \mbox{XProtoNet} to that of two recent automated diagnosis methods~\cite{li2018thoracic,liu2019align} using bounding box supervision, which use \mbox{ResNet-50}~\cite{he2016deep} as the backbone.
Table~\ref{tab:classification_cv} shows the performances based on a five-fold cross-validation.
Despite having no additional supervision, \mbox{XProtoNet} achieves the best performance for most diseases.

\subsection{XProtoNet with Prior Condition}\label{sec:additional_supervision}
\vspace{-2pt}
As XProtoNet provides predictions based on prototypes that are exposed explicitly, we can instruct it to diagnose using specific signs of diseases by forcing the prototypes to present those signs. We conduct analysis with the prior condition that the prototypes of \mbox{XProtoNet} should present the features within the bounding box annotations.

XProtoNet is trained with both box-annotated and box-unannotated data. We set both $\lambda_\text{clst}$ and $\lambda_\text{sep}$ to 1.5 for the box-annotated data and both to 0.5 for the box-unannotated data.
To utilize the bounding box annotations during training, we extract the feature vectors from the feature maps within the bounding boxes as $\textbf{f}_{\pveckc}^\text{bbox}(\xvec)=\sum_{u\in\text{bbox}}{M_{\pveckc,u}(\xvec) F_u(\xvec)}$, where bbox denotes the spatial location inside the bounding box. We also change $L_1$ loss on the occurrence map for the box-annotated data to $\sum_{\pveckc\in\mathcal{P}^c}{\sum_{u\not\in\text{bbox}}{M_{\pvec_k^c,u}(\xvec)}}$ to suppress the area outside the bounding box from being activated in the occurrence map.
To enable the prototypes to present the features within the bounding boxes, the prototype vectors are replaced with their most similar feature vectors $\textbf{f}_{\pveckc}^\text{bbox}$ from the box-annotated data, instead of the feature vectors $\textbf{f}_{\pveckc}$ from the box-unannotated data.

\begin{figure}[t]
	\centering
    \includegraphics[width=0.79\columnwidth]{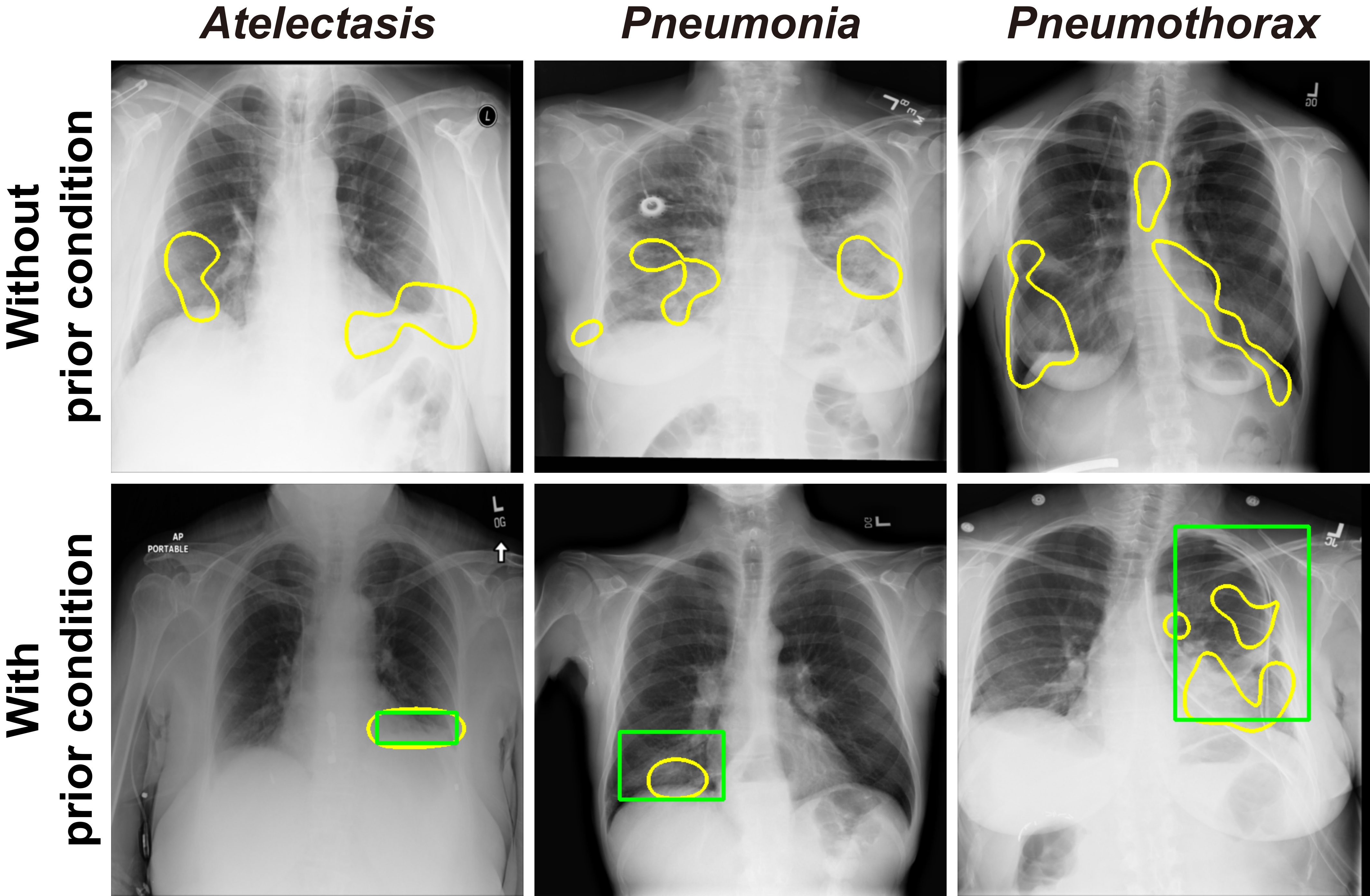}
    \caption{Examples of the learned prototypes of XProtoNet trained with and without the prior condition. Yellow contours denote the learned prototypes and green boxes denote the ground truth bounding boxes from the dataset.}
    \label{fig:supervision}
\vspace{-5pt}
\end{figure}

Figure~\ref{fig:supervision} shows the learned prototypes of XProtoNet trained with and without the prior condition. Owing to the constraint, the prototypes of XProtoNet trained with the prior condition present disease-representative features within the bounding box annotations. Although this can be a strong constraint for the model, there is no significant difference in the diagnostic performance: the mean AUC scores over 14 diseases of XProtoNet trained with and without the prior condition are 0.850 and 0.849, respectively.
Therefore, using the prior condition, we enable XProtoNet diagnoses based on the specific features, thus rendering the system more trustworthy.

\section{Conclusion}
\vspace{-3pt}
XProtoNet is an automated diagnostic framework for chest radiography that ensures human interpretability as well as high performance. XProtoNet can provide not only a local explanation for a given X-ray image but also a global explanation for each disease, which is not provided by other diagnostic methods. Despite the constraints imposed by the interpretability requirement, it achieves state-of-the-art diagnostic performance by predicting the dynamic areas where disease-representative features may be found.

With a post-hoc explanation such as localization, it is difficult to understand how a model classifies an input image. XProtoNet is one of only a very few attempts to design an explicitly interpretable model. Further research on interpretable systems using DNNs will therefore encourage the trustworthiness of the automated diagnosis system.

\bigskip
\vspace{-1pt}
\noindent\textbf{Acknowledgements:}
This work was supported by the National Research Foundation of Korea (NRF) grant funded by the Korea government (Ministry of Science and ICT) [2018R1A2B3001628], AIR Lab (AI Research Lab) in Hyundai \& Kia Motor Company through HKMC-SNU AI Consortium Fund, and the BK21 FOUR program of the Education and Research Program for Future ICT Pioneers, Seoul National University in 2021.

{\small
\bibliographystyle{ieee_fullname}
\bibliography{egbib}
}

\end{document}